# A Two-Round Variant of EM for Gaussian Mixtures

**Sanjoy Dasgupta**[*]
AT&T Labs – Research

**Leonard J. Schulman**
Georgia Institute of Technology

## Abstract

We show that, given data from a mixture of $k$ well-separated spherical Gaussians in $\mathbb{R}^n$, a simple two-round variant of EM will, with high probability, learn the centers of the Gaussians to near-optimal precision, if the dimension is high ($n \gg \log k$). We relate this to previous theoretical and empirical work on the EM algorithm.

## 1 Introduction

At present EM is the method of choice for learning mixtures of Gaussians. A series of theoretical and experimental studies over the past three decades have contributed to the collective intuition about this algorithm. We will reinterpret a few of these results in the context of a new performance guarantee.

A standard criticism of EM is that it converges very slowly. Simulations performed by Redner and Walker (1984), and others, demonstrate this decisively for one-dimensional mixtures of two Gaussians. It is also known that given data from a mixture of Gaussians, when EM gets close to the true solution, it exhibits *first-order convergence*. Roughly speaking, the idea is this: given $m$ data points from a mixture with parameters (means, covariances, and mixing weights) $\theta^*$, where $m$ is very large, the log-likelihood has a local maximum at some set of parameters $\theta^m$ close to $\theta^*$. Let $\theta^{(t)}$ denote EM's parameter-estimates at time $t$. It can be shown (cf. Taylor expansion) that when $\theta^{(t)}$ is near $\theta^m$,

$$\|\theta^{(t+1)} - \theta^m\| \leq \lambda \cdot \|\theta^{(t)} - \theta^m\|,$$

where $\lambda \in [0, 1)$ and $\|\cdot\|$ is some norm.[1] If the Gaussians are closely packed then $\lambda$ is close to one; if they are very far from one another then $\lambda$ is close to zero. These results are the work of many researchers; a summary can be found in the overview paper of Redner and Walker (1984).

Xu and Jordan (1995) present theoretical results which mitigate some of the pessimism of first-order convergence, particularly in the case of well-separated mixtures, and they note that moreover near-optimal log-likelihood is typically reached in just a few iterations. We also argue in favor of EM, but in a different way. We ask, how close does $\theta^{(t)}$ have to be to $\theta^m$ for slow convergence to hold? Let $d(\theta_1, \theta_2)$ denote the maximum Euclidean distance between the respective means of $\theta_1$ and $\theta_2$. For one-dimensional data, it can be seen quite easily from canonical experiments (Redner and Walker, 1984) that convergence is slow even if $d(\theta^{(t)}, \theta^m)$ is large. However, our results suggest that this no longer holds in higher dimension. For reasonably well-separated spherical Gaussians in $\mathbb{R}^n$ (where *separation* is defined precisely in the next section), convergence is very fast until $d(\theta^{(t)}, \theta^m) \approx e^{-\Omega(n)}$. In fact, we can make EM attain this accuracy in just two rounds. The error $e^{-\Omega(n)}$ is so miniscule for large $n$ that subsequent improvements are not especially important.

Practitioners have long known that if the data has $k$ clusters, then EM should be started with more than $k$ centers, and these should at some stage be pruned. We present a simple example to demonstrate exactly why this is necessary, and obtain an expression for the number of initial centers which should be used: $O(\frac{1}{w_{min}} \log k)$, where $w_{min}$ is a lower bound on the smallest mixing weight. The typical method of pruning is to remove Gaussian-estimates with very low mixing weight (known as *starved clusters*). Our theoretical analysis shows that this is not enough, that there is another type of Gaussian-estimate, easy to detect, which also needs to be pruned. Specifically, it is possible (and frequently occurs in simulations) that two of EM's Gaussian-estimates share the same cluster, each with relatively high mixing weight. We present a very simple, provably correct method of detecting this situation and correcting it.

It is widely recognized that a crucial issue in the performance of EM is the choice of initial parameters. For the means, we use the popular technique of picking initial center-estimates randomly from the data set. This is shown to be adequate for the performance guarantee we derive. Our analysis also makes it clear that it is vitally important to pick good initial estimates of the covariances, a subject which has received somewhat less attention. We use a clever initializer whose origin we are unable to trace but which is mentioned in Bishop's text (1995).

---

[*]Work done while at University of California, Berkeley.
[1]This might not seem so bad, but contrast it with *second-order convergence*, in which $\|\theta^{(t+1)} - \theta^m\| \leq \lambda \cdot \|\theta^{(t)} - \theta^m\|^2$.

Our central performance guarantee requires that the clusters actually look spherical-Gaussian, more specifically that the data points are drawn i.i.d. from some (unknown) mixture of spherical Gaussians. We show that if the clusters are reasonably well-separated (in a precise sense), and if the dimension $n \gg \log k$ then only two rounds of EM are required to learn the mixture to within near-optimal precision, with high probability $1 - k^{-\Omega(1)}$. Our measure of accuracy is the function $d(\cdot, \cdot)$ introduced above. The precise statement of the theorem can be found in Section 3.4, and applies not only to EM but also to other similar schemes, including for instance some of the variants of EM and $k$-means introduced by Kearns, Mansour, and Ng (1997).

Performance guarantees for clustering will inevitably involve some notion of the *separation* between different clusters. There are at least two natural ways of defining this. Take for simplicity the case of two $n$-dimensional Gaussians $N(\mu_1, I_n)$ and $N(\mu_2, I_n)$. If each coordinate (attribute) provides a little bit of discriminative information between the two clusters, then on each coordinate the means $\mu_1$ and $\mu_2$ differ by at least some small amount, say $\delta$. The $L_2$ distance between $\mu_1$ and $\mu_2$ is then at least $\delta\sqrt{n}$. As further attributes are added, the distance between the centers grows, and the two clusters become more clearly distinguishable from one another. This is the usual rationale for using high-dimensional data: the higher the dimension, the easier (in an information-theoretic sense) clustering should be. The only problem then, is whether there are algorithms which can efficiently exploit the tradeoff between this high information content and the curse of dimensionality. This viewpoint suggests that the $L_2$ distance between the centers of $n$-dimensional clusters can reasonably be measured in units of $\sqrt{n}$, and that it is most important to develop algorithms which work well under the assumption that this distance is some constant times $\sqrt{n}$. On the other hand, it should be pointed out that if $\|\mu_1 - \mu_2\| = \delta\sqrt{n}$ for some constant $\delta > 0$, then for large $n$ the overlap in probability mass between the two Gaussians is miniscule, exponentially small in $n$. Therefore, it should not only be interesting but also possible to develop algorithms which work well when the $L_2$ distance between centers of clusters is much smaller, for instance some constant independent of the dimension (as opposed to $O(\sqrt{n})$).

Where do EM's requirements fall in this spectrum of separation? We show that EM works well in at least a large part of this span, when the distance between clusters is bigger than $n^{1/4}$.

In the final section of the paper, we discuss a crucial issue: what features of our main assumption (that the clusters are high-dimensional Gaussians) make such a strong statement about EM possible? This assumption is also the basis of all the other theoretical results mentioned above, but can real data sets reasonably be expected to satisfy it? If not, in what way can it usefully be relaxed?

## 2 High-dimensional Gaussians

A spherical Gaussian $N(\mu, \sigma^2 I_n)$ assigns to point $x \in \mathbb{R}^n$ the density

$$p(x) = \frac{1}{(2\pi)^{n/2}\sigma^n} \exp\left(-\frac{\|x-\mu\|^2}{2\sigma^2}\right),$$

$\|\cdot\|$ being Euclidean distance. If $X = (X_1, \ldots, X_n)$ is randomly chosen from $N(0, \sigma^2 I_n)$ then its coordinates are i.i.d. $N(0, \sigma^2)$ random variables. Each coordinate has expected squared value $\sigma^2$ so $\mathbf{E}\|X\|^2 = \mathbf{E}(X_1^2 + \cdots + X_n^2) = n\sigma^2$. It then follows by a large deviation bound that $\|X\|^2$ will be tightly concentrated around $n\sigma^2$:

$$\mathbf{P}(|\|X\|^2 - n\sigma^2| > \epsilon n\sigma^2) \leq e^{-n\epsilon^2/24}.$$

This bound and others like it will be discussed in Section 4. It means that almost the entire probability mass of $N(0, \sigma^2 I_n)$ lies in a thin shell at a radius of $\sigma\sqrt{n}$ from the origin. This does not contradict the fact that the density of the Gaussian is highest at the origin, since the surface area at distance $r$ from the origin, $0 \leq r \leq \sigma\sqrt{n}$, increases faster than the density at distance $r$ decreases (Bishop, 1995, exercise 1.4).

It is natural therefore to think of a Gaussian $N(\mu, \sigma^2 I_n)$ as having *radius* $\sigma\sqrt{n}$. We say two Gaussians $N(\mu_1, \sigma_1^2 I_n), N(\mu_2, \sigma_2^2 I_n)$ in $\mathbb{R}^n$ are *c-separated* if

$$\|\mu_1 - \mu_2\| \geq c \max\{\sigma_1, \sigma_2\}\sqrt{n},$$

that is, if they are $c$ radii apart (Dasgupta, 1999). A mixture of Gaussians is $c$-separated if the Gaussians in it are pairwise $c$-separated. In general we will let $c_{ij}$ denote the separation between the $i^{th}$ and $j^{th}$ Gaussians, and $c = \min_{i \neq j} c_{ij}$. We can reasonably expect that the difficulty of learning a mixture of Gaussians increases as $c$ decreases. For non-spherical Gaussians this definition can be extended readily by thinking of the radius of $N(\mu, \Sigma)$ as being $\sqrt{\text{trace}(\Sigma)}$.

A 2-separated mixture contains clusters with almost no overlap. In $\mathbb{R}^n$ for large $n$, this is true even of a $\frac{1}{100}$-separated mixture, because for instance, two spheres of radius $\sqrt{n}$ with centers $\frac{1}{100}\sqrt{n}$ apart share only a tiny fraction of their volume. One useful way of thinking about a pair of $c$-separated Gaussians is to imagine that on each coordinate their means differ by $c$. If $c$ is small, then the projection of the mixture onto any one coordinate will look unimodal. This might also be true of a projection onto a few coordinates. But for large $n$, when all coordinates are considered together, the distribution will cease to look unimodal. This is precisely the reason for using high-dimensional data.

What values of $c$ can be expected of real-world data sets? This will vary from case to case. As an example, we analyzed a canonical data set consisting of handwritten digits collected by USPS. Each digit was represented as a vector in $[-1, 1]^{256}$. We fit a mixture of ten (non-spherical) Gaussians to this data set, by doing each digit separately, and found that it was 0.63-separated.

## 3 A two-round variant of EM: the case of common covariance

It is instructive and convenient to start with the subcase in which data is drawn from a mixture of $k$ Gaussians with the same spherical covariance matrix $\sigma^2 I_n$, for some unknown $\sigma^2$. We will show that if $n \gg \log k$, EM can be made to work well in just two rounds.

### 3.1 The EM algorithm

Given a data set $S \subset \mathbb{R}^n$, the EM algorithm (for a mixture of $k$ Gaussians with common spherical covariance) works by first choosing starting values $\mu_i^{\langle 0 \rangle}, w_i^{\langle 0 \rangle}, \sigma^{\langle 0 \rangle}$ for the parameters, and then updating them iteratively according to the following two-step procedure (at time $t$).

**E step** Let $\tau_i \sim N(\mu_i^{\langle t \rangle}, \sigma^{\langle t \rangle 2} I_n)$ denote the density of the $i^{th}$ Gaussian-estimate. For each data point $x \in S$, and each $1 \leq i \leq k$, compute

$$p_i^{\langle t+1 \rangle}(x) = \frac{w_i^{\langle t \rangle} \tau_i(x)}{\sum_j w_j^{\langle t \rangle} \tau_j(x)},$$

the conditional probability that $x$ comes from the $i^{th}$ Gaussian with respect to the current parameters.

**M step** Now update the various parameters in an intuitive way. Denote the size of $S$ by $m$.

$$
\begin{aligned}
w_i^{\langle t+1 \rangle} &= \frac{1}{m} \sum_{x \in S} p_i^{\langle t+1 \rangle}(x) \\
\mu_i^{\langle t+1 \rangle} &= \frac{\sum_{x \in S} x \, p_i^{\langle t+1 \rangle}(x)}{m w_i^{\langle t+1 \rangle}} \\
\sigma^{\langle t+1 \rangle 2} &= \frac{1}{mn} \sum_{x \in S} \sum_{i=1}^{k} \|x - \mu_i^{\langle t+1 \rangle}\|^2 \, p_i^{\langle t+1 \rangle}(x)
\end{aligned}
$$

### 3.2 The main issues

It will turn out that when the separation of a mixture in $\mathbb{R}^n$ is $c \gg n^{-1/4}$ then the chance that two points from differ-ent Gaussians are closer together than two points from the same Gaussian, is tiny, $e^{-\Omega(poly(n))}$. Therefore an examination of interpoint distances is enough to almost perfectly cluster the data. A variety of different algorithms will work well under these circumstances, and EM is no exception.

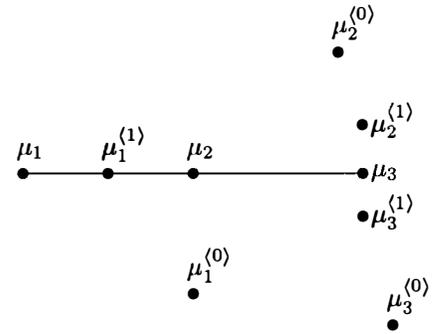

Figure 1: For this mixture, the positions of the center-estimates do not move much after the first step of EM.

Suppose the true number of Gaussians, $k$, is known. Let $S$ denote the entire data set, and $S_i$ the points drawn from the $i^{th}$ true Gaussian $N(\mu_i, \sigma^2 I_n)$. A common way to initialize EM is to pick $l$ data points at random from $S$, and to use these as initial *center-estimates* $\mu_i^{\langle 0 \rangle}$. How large should $l$ be? It turns out that if these $l$ points include at least one point from each $S_i$, then EM can be made to perform well. This suggests $l = \Omega(k \log k)$. Conversely, if the initial centers miss some $S_i$, then EM might perform poorly.

Here is a concrete example (Figure 1). Let $n$ denote some high dimension, and place the $k$ true Gaussians $N(\mu_1, I_n), \ldots, N(\mu_k, I_n)$ side by side in a line, leaving a distance of $3\sqrt{n}$ between consecutive Gaussians. Assign them equal mixing weights. As before let $S_i$ be the data points from the $i^{th}$ Gaussian, and choose EM's initial center-estimates from the data. Suppose the initial centers contain nothing from $S_1$, one point from $S_2$, and at least one point from $S_3$. The probability of this event is at least some constant. Then no matter how long EM is run, it will assign just one Gaussian-estimate to the first two true Gaussians. In the first round of EM, the point from $S_2$ (call it $\mu_1^{\langle 0 \rangle}$) will move between $\mu_1$ and $\mu_2$. It will stay there, right between the two true centers. None of the other center-estimates $\mu_i^{\langle t \rangle}$ will ever come closer to $\mu_2$; their distance from it is so large that their influence is overwhelmed by that of $\mu_1^{\langle t \rangle}$. This argument can be formalized easily using the large deviation bounds of the next section.

How about the initial choice of variance? When the Gaussians have a common spherical covariance, this is not all that important, except that a huge overestimate might cause slower convergence. We will use a fairly precise estimator, a variant of which is mentioned in Bishop's text (1995).

After one round of EM, the center-estimates are pruned to leave exactly one per true Gaussian. This is accomplished in a simple manner. First, remove any center-estimates with very low mixing weight (this is often called "cluster starvation"). Any remaining center-estimate (originally chosen, say, from $S_i$) has relatively high mixing weight, and we can show that as a result of the first EM iteration, it will have moved close to $\mu_i$. A trivial clustering heuristic, due to Hochbaum and Shmoys (1985), is then good enough to select one center-estimate near each $\mu_i$.

With exactly one center-estimate per (true) Gaussian, a second iteration of EM will accurately retrieve the means, covariance, and mixing weights. In fact the clustering of the data (the fractional labels assigned by EM) will be almost perfect, that is to say, each fractional label will be close to zero or one, and will in almost all cases correctly identify the generating Gaussian. Therefore further iterations will not help much: these additional iterations will move the center-estimates around by at most $e^{-\Omega(n)}$.

### 3.3 The simplified algorithm

Here is a summary of the modified algorithm, given $m$ data points in $\mathbb{R}^n$ which have been generated by a mixture of $k$ Gaussians. The value of $l$ will be specified later; for the time being it can be thought of as $O(k \log k)$.

**Initialization** Pick $l$ data points at random as starting estimates $\mu_i^{\langle 0 \rangle}$ for the Gaussian centers. Assign them identical mixing weights $w_i^{\langle 0 \rangle} = \frac{1}{l}$. For an initial estimate of variance use
$$\sigma^{\langle 0 \rangle 2} = \frac{1}{2n} \min_{i \neq j} \| \mu_i^{\langle 0 \rangle} - \mu_j^{\langle 0 \rangle} \|^2.$$

**EM** Run one round of EM. This yields modified estimates $\mu_i^{\langle 1 \rangle}, \sigma^{\langle 1 \rangle}, w_i^{\langle 1 \rangle}$.

**Pruning** Remove all center-estimates whose mixing weights are below $w_T = \frac{1}{2l} + \frac{2}{m}$. Prune the remaining center-estimates down to just $k$ of them:

- Compute distances between center-estimates.
- Choose one of these centers arbitrarily.
- Pick the remaining $k - 1$ iteratively as follows: pick the center farthest from the ones picked so far. (The distance from a point $x$ to a set $S$ is $\min_{y \in S} \|x - y\|$, where $\|\cdot\|$ is the $L_2$ norm.)

Call the resulting center-estimates $\tilde{\mu}_i^{\langle 1 \rangle}$ (where $1 \leq i \leq k$). Set the mixing weights to $\tilde{w}_i^{\langle 1 \rangle} = \frac{1}{k}$ and the standard deviation to $\tilde{\sigma}^{\langle 1 \rangle} = \sigma^{\langle 0 \rangle}$.

**EM** Run one more step of EM, starting at the $\{\tilde{\mu}_i^{\langle 1 \rangle}, \tilde{w}_i^{\langle 1 \rangle}, \tilde{\sigma}^{\langle 1 \rangle}\}$ parameters and yielding the final estimates $\mu_i^{\langle 2 \rangle}, w_i^{\langle 2 \rangle}, \sigma^{\langle 2 \rangle}$.

### 3.4 The main result

Now that the notation and algorithm have been introduced, we can state the main theorem for the case of common covariances; a similar result holds when the Gaussians have different spherical covariance matrices (Section 8).

**Theorem 1** *Say $m$ data points are generated from a c-separated mixture of $k$ Gaussians $w_1 N(\mu_1, \sigma^2 I_n) + \cdots + w_k N(\mu_k, \sigma^2 I_n)$ in $\mathbb{R}^n$. Let $S_i$ denote the points from the $i^{th}$ Gaussian, and let $w_{min} = \min_i w_i$. Further, define*

$$\alpha = \frac{1}{2} - \frac{\ln 30 \max(1, c^{-2})}{\ln n} \quad \text{and} \quad \beta = \frac{c^2 n}{512 \ln m}.$$

*Then, assuming $\alpha > 0$ and $\min(n, c^2 n) \geq 18 + 8 \ln n$ and $m \geq \max(4l^2, 2^{18} c^{-4})$, with probability at least $1 - m^2 e^{-\Omega(n^{2\alpha})} - k e^{-\Omega(l w_{min})} - m^{-(\beta - 1)}$ the variant of EM described above will produce final center-estimates which satisfy*

$$\|\mu_i^{\langle 2 \rangle} - \mu_i\| \leq \|mean(S_i) - \mu_i\| + e^{-\Omega(c^2 n)}.$$

The proof of this theorem will be sketched over the next four sections; the details can be found in the full version of the paper. A few words of explanation are in order at this stage. First of all, the constants mentioned in the theorem should not be a source of concern since no attempt has been made to optimize them. Second, the best that can be hoped is that $\mu_i^{\langle 2 \rangle} = \text{mean}(S_i)$; therefore, the final error bound on the center-estimates is very close to optimal. Finally notice that $\alpha > 0$ requires that $c \gg n^{-1/4}$, and that in order to make the probability of failure at most $k^{-\Omega(1)}$, it is necessary to set $l = O(\frac{1}{w_{min}} \log k)$, to use $m = l^2 \text{poly}(k)$ samples, and to assume that $n^{2\alpha} = \Omega(\log k)$.

## 4 Initialization

We will show that the two-round algorithm retrieves the true Gaussians with high probability. This result hinges crucially upon large deviation bounds for the lengths of points drawn from a Gaussian (Dasgupta, 1999, Lemma 14).

**Lemma 2** *Pick $X$ from $N(0, I_n)$. For any $\epsilon \in (0, 1)$,*

$$\mathbf{P}(|\|X\|^2 - n| \geq \epsilon n) \leq 2 e^{-n \epsilon^2 / 24}.$$

*Thus for any $\alpha > 0$, $\|X\|^2 \in [n - n^{1/2+\alpha}, n + n^{1/2+\alpha}]$ with probability at least $1 - 2 e^{-n^{2\alpha}/24}$.*

It can similarly be shown that the distance between two points from the same Gaussian (or from different Gaussians) is sharply concentrated around its expected value.

**Lemma 3** *If $X$ is chosen from $N(\mu_i, \sigma_i^2 I_n)$ and $Y$ is chosen independently from $N(\mu_j, \sigma_j^2 I_n)$ then for any $\alpha > 0$,* *the chance that $\|X - Y\|^2$ does not lie in the range $\|\mu_i - \mu_j\|^2 + (\sigma_i^2 + \sigma_j^2)(n \pm n^{1/2+\alpha}) \pm 2\|\mu_i - \mu_j\|\sqrt{\sigma_i^2 + \sigma_j^2} \cdot n^\alpha$ is at most $2e^{-n^{2\alpha}/24} + e^{-n^{2\alpha}/2}$.*

**Corollary 4** *Draw $m$ data points from a c-separated mixture of $k$ Gaussians with common covariance matrix $\sigma^2 I_n$ and smallest mixing weight at least $w_{min}$. Let $S_i$ denote the points from the $i^{th}$ Gaussian. Then for any $\alpha > 0$, with probability at least $1 - (m^2 + 2km)e^{-n^{2\alpha}/24} - ke^{-mw_{min}/32} - \frac{1}{2}m^2 e^{-n^{2\alpha}/2} - kme^{-n^{2\alpha}/2}$,*

*(1) for any $x, y \in S_j$, $\|x - y\|^2 = 2\sigma^2 n \pm 2\sigma^2 n^{1/2+\alpha}$;*

*(2) for $x \in S_i, y \in S_j, i \neq j$, $\|x - y\|^2 = (2 + c_{ij}^2)\sigma^2 n \pm (2 + 2\sqrt{2}c_{ij})\sigma^2 n^{1/2+\alpha}$;*

*(3) for any data point $y \in S_j$, $\|y - \mu_j\|^2 = \sigma^2 n \pm \sigma^2 n^{1/2+\alpha}$ while for $i \neq j$, $\|y - \mu_i\|^2 = (1 + c_{ij}^2)\sigma^2 n \pm (1 + 2c_{ij})\sigma^2 n^{1/2+\alpha}$; and*

*(4) each $|S_i| \geq \frac{3}{4}mw_i$.*

This means that if the mixture is $c$-separated, then points from the same Gaussian are at squared distance about $2\sigma^2 n$ from each other while points from different Gaussians are at squared distance at least about $2(1 + \Omega(c^2))\sigma^2 n$ from each other. The standard deviation of these estimates is around $\sigma^2 n^{1/2}$. If $c^2 n \gg n^{1/2}$ then this standard deviation will be overwhelmed by the separation between clusters, and therefore points from the same cluster will almost always be closer together than points from different clusters. In such a situation, interpoint distances will reveal enough information for clustering and it should, in particular, be possible to make EM work well. We first establish some simple guarantees about the initial conditions.

**Lemma 5** *If $l > k$ and each $w_i \geq w_{min}$ then with probability at least $1 - ke^{-lw_{min}} - ke^{lw_{min}/48}$,*

*(a) every Gaussian is represented in the initial center-estimates;*

*(b) the $i^{th}$ Gaussian provides at most $\frac{5}{4}lw_i$ initial center-estimates, for all $1 \leq i \leq k$; and*

*(c) $\sigma^{(0)2} = \sigma^2(1 \pm n^{-1/2+\alpha})$.*

**Remark** All the theorems of the following sections are made under the additional hypothesis that Corollary 4 and Lemma 5 hold, for some fixed $\alpha \in (0, \frac{1}{2})$.

## 5 The first round of EM

What happens during the first round of EM? The first thing we clarify is that although in principle EM allows "soft" assignments in which each data point is fractionally distributed over various clusters, in practice for large $n$ every data point will give almost its entire weight to center-estimates from one (true) cluster. This is because in high dimension, the distances between clusters are so great that there is just a very narrow region between two clusters where there is any ambiguity of assignment, and the probability that points fall within this region is miniscule.

Recall that we are defining $S_i$ as the data points drawn from the true Gaussian $N(\mu_i, \sigma^2 I_n)$. Combining the last few lemmas tells us that if $c^2 n \gg \ln l$, in the first round of EM each data point in $S_i$ will have almost all its weight assigned to center-estimates $\mu_j^{(0)}$ in $S_i$. Therefore, fix attention on a specific Gaussian, say $N(\mu_1, \sigma^2 I_n)$. Without loss of generality, $\mu_1 = 0$ and the initial center-estimates $\mu_1^{(0)}, \ldots, \mu_q^{(0)}$ came from this Gaussian, that is, they are in $S_1$. We know from Lemma 5 that $1 \leq q \leq \frac{5}{4}lw_1$.

Say that center-estimate $\mu_1^{(0)}$ receives a reasonably high mixing weight after the first round, specifically that $w_1^{(1)} \geq w_T$ (by a lemma of the next section, at least one of $\mu_1^{(0)}, \ldots, \mu_q^{(0)}$ must have this property). We will show that its new value $\mu_1^{(1)}$ is much closer to $\mu_1$ (that is, to the origin). For any data point $x \in S$, let $p_i(x)$ denote the (fractional) weight that $x$ gives to $\mu_i^{(0)}$ during the first round of EM. Then

$$\mu_1^{(1)} = \frac{\sum_{x \in S} p_1(x) x}{\sum_{x \in S} p_1(x)}.$$

By our previous discussion, the most important contribution here is from points $x$ in $S_1$. So let's ignore other terms for the time being and focus upon the central quantity

$$\mu_1^* = \frac{\sum_{x \in S_1} p_1^*(x) x}{\sum_{x \in S_1} p_1^*(x)}.$$

where $p_1^*(x)$ is the fractional weight assigned to $x$ assuming no centers other than $\mu_1^{(0)}, \ldots, \mu_q^{(0)}$ are active, that is,

$$p_1^*(x) = \frac{p_1(x)}{p_1(x) + \cdots + p_q(x)}.$$

We have already asserted that the total mixing weight assigned to $\mu_1^{(0)}$, namely $\sum_{x \in S} p_1(x) \approx \sum_{x \in S_1} p_1^*(x)$, is quite high. How can we bound $\|\mu_1^* - \mu_1\|$? The first step is to notice that when the data points in $S_1$ are being assigned to centers $\mu_j^{(0)}, j = 1, \ldots, q$, the fractional assignments $p_j^*(\cdot)$ can be made entirely on the basis of the projections of these points into the subspace spanned by $\mu_1^{(0)}, \ldots, \mu_q^{(0)}$ (since the Gaussian-estimates have a common, and spherically symmetric, covariance). Specifically, let $L$ denote this subspace, which has some dimension $d \leq q$ (and of course $d \leq n$). Rotate the axes so that $L$ coincides with the first $d$ coordinates. Write each point $X \in \mathbb{R}^n$ in the form $(X_L, X_R)$. Note that $\mu_1^{(0)}, \ldots, \mu_q^{(0)}$ have zeros in their last $n - d$ coordinates.

Each data point $X \in S_1$ is chosen from $N(0, \sigma^2 I_n)$ (recall we are assuming $\mu_1 = 0$ for convenience) and then divided between the various center-estimates. We can replace the process

- Pick $X$ according to $N(0, \sigma^2 I_n)$.
- Divide it between $\mu_1^{\langle 0 \rangle}, \ldots, \mu_q^{\langle 0 \rangle}$.

by the process

- Pick $X_L$ according to $N(0, \sigma^2 I_d)$.
- Divide it between $\mu_1^{\langle 0 \rangle}, \ldots, \mu_q^{\langle 0 \rangle}$.
- Now pick $X_R$ according to $N(0, \sigma^2 I_{n-d})$.

Then

$$\mu_1^* = \frac{\sum_{x \in S_1} p_1^*(x) x_L}{\sum_{x \in S_1} p_1^*(x)} + \frac{\sum_{x \in S_1} p_1^*(x) x_R}{\sum_{x \in S_1} p_1^*(x)}.$$

The last term is easy to bound because, even conditional upon $p_1^*(x)$, the $x_R$ look like random draws from $N(0, \sigma^2 I_{n-d})$. The other is more difficult because the $x_L$ are not independent of the $p_1^*(x)$. A simple estimate is to use the fact that each $\|x_L\|$ is about $O(\sqrt{d})$; therefore a convex combination of $x_L$'s will have length at most about $O(\sqrt{d}) \leq O(\sqrt{q})$. This works well when $q$ is very small; by a more careful analysis we will now arrive at a bound of $O(\sqrt{\log q})$.

The main thing working in our favor is that $\sum_{x \in S_1} p_1^*(x)$ is not too small. Say this value is $r$. Suppose no fractional assignments were allowed. Then we would know that $r$ whole data points were assigned to $\mu_1^{\langle 0 \rangle}$, and it would be enough to prove that *any* $r$ points out of $S_1$ average to something fairly close to the origin.

However, fractional assignments are allowed, so we must remove this annoyance somehow.

**Lemma 6** *Given fractional labels $f(y) \in [0, 1]$ for a finite set of points $y \in \mathbb{R}^d$, there is a corresponding set of binary labels $g(y) \in \{0, 1\}$ such that $1 + \sum_y g(y) \geq \sum_y f(y)$ and $\left\| \frac{\sum_y g(y)y}{\sum_y g(y)} \right\| \geq \left\| \frac{\sum_y f(y)y}{\sum_y f(y)} \right\|$.*

*Proof.* Let $A$ denote $(\sum_y f(y)y)/(\sum_y f(y))$. Suppose for convenience that $A$ lies along some coordinate axis, say the positive $z$ axis. Consider the hyperplane $z = \|A\|$. Divide the $y$'s into two sets: the points $Y_<$ which lie in the half-space $z < \|A\|$ and the points $Y_\geq$ which lie in the half-space $z \geq \|A\|$. We will adjust the weights of points according to which side of the hyperplane they lie on. In general, we do not mind increasing the weights of points in $Y_\geq$ and decreasing the weights of those in $Y_<$ because this will guarantee that the resulting weighted average is in the half-space $z \geq \|A\|$ and is therefore further from the origin than $A$. The only problem is that we are allowed to reduce the overall weight by at most one.

The new weights $g(y)$ are assigned according to the following procedure:

- Set all $g(y) = f(y)$.

- For each point $y \in Y_\geq$, increase its weight to $g(y) = 1$. This increases the overall weight $\sum_y g(y)$ and ensures that the resulting convex combination lies in the half-space $z \geq \|A\|$.

- Consider the points $y \in Y_<$. Out of them, pick (1) the point $u$ closest to the hyperplane $z = \|A\|$ (ie. with the highest $z$ coordinate) and which has weight $g(u) < 1$ and (2) the point $v$ farthest from the hyperplane (with the smallest $z$ coordinate) and which has weight $g(v) > 0$. Increase the weight of $u$ by $\min(g(v), 1 - g(u))$ and decrease the weight of $v$ by this same amount. Each such adjustment does not alter the overall weight $\sum_y g(y)$ and drives the $z$ coordinate of $(\sum_{y \in Y_<} g(y)y)/(\sum_{y \in Y} g(y))$ closer to $\|A\|$. Iterate this process until there remains at most one point with a fractional weight; at most $|Y_<|$ iterations are needed. Remove this last point.

This procedure guarantees that $\sum_y g(y) \geq (\sum_y f(y)) - 1$ and that $(\sum_y g(y)y)/(\sum_y g(y))$ lies in the half-space $z \geq \|A\|$. Therefore its norm must be at least $\|A\|$. ∎

Next we show that there is no large subset of $S_1$ whose average has very large norm (we are still assuming $\mu_1 = 0$).

**Lemma 7** *Pick $|S_1|$ points randomly from $N(0, I_d)$. Choose any $\beta > 0$. Then with probability at least $1 - m^{-\beta}$, for any $v \geq \max(\beta, d)$, there is no subset of $S_1$ of size $\geq v$ whose average has squared length more than $4(\ln 2e|S_1|/v + (\beta/d) \ln m)$.*

These last two lemmas can be used to bound the contribution of the $x_L$'s to $\mu_1^*$. The $x_R$'s are independent of the $p_1^*(x)$'s; therefore their contribution is easy to analyze. Putting these together yields the next lemma.

**Lemma 8** *Choose any $\beta > 0$. If $\sum_{x \in S_1} p_1^*(x) \geq r + 1$, where $r \geq \max(\beta, d)$ then with probability at least $1 - m^{-\beta} - e^{-n/8}$,*

$$\|\mu_1^*\|^2 \leq 4\sigma^2 \left( \ln \frac{2e|S_1|}{r} + \frac{\beta}{d} \ln m \right) + \frac{2\sigma^2 n}{r + 1}.$$

*Proof.* Let $f(x) = p_1^*(x)$ be the (fractional) weight with which $x \in S_1$ is assigned to $\mu_1^{\langle 0 \rangle}$. Obtain the binary weights $g(\cdot)$ as in Lemma 6; therefore $\sum_{x \in S_1} g(x) \geq r$. As before, divide the coordinates into two groups, $L$ and $R$. We will consider the averages $A_L$ and $A_R$ of these two parts separately. By Lemmas 6 and 7, with probability at least $1 - m^{-\beta}$,

$$\|A_L\|^2 = \left\|\frac{\sum_{x \in S_1} f(x)x_L}{\sum_{x \in S_1} f(x)}\right\|^2 \leq \left\|\frac{\sum_{x \in S_1} g(x)x_L}{\sum_{x \in S_1} g(x)}\right\|^2$$

$$\leq 4\sigma^2 \left(\ln \frac{2e|S_1|}{r} + \frac{\beta}{d} \ln m\right).$$

For $A_R$, if $d = n$ then $A_R = 0$ and we have nothing to worry about. If $d < n$, write $n - d = \gamma n$ (where $\gamma \in [\frac{1}{n}, 1]$), and

$$A_R = \frac{\sum_{x \in S_1} f(x)x_R}{\sum_{x \in S_1} f(x)} \stackrel{d}{=} N(0, \frac{\sigma^2}{t} I_{\gamma n}),$$

where $t = (\sum_x f(x))^2/(\sum_x f(x)^2) \geq \sum_x f(x)$ (since $f(x) \geq f(x)^2$) and so $t \geq r + 1$. The chance that a $N(0, I_{\gamma n})$ random variable has squared length more than $2n$ is at most $e^{-n/8}$. Therefore $\|A_R\|^2 \leq 2\sigma^2 n/(r+1)$ with probability at least $1 - e^{-n/8}$. To finish the lemma note that $\mu_1^* = (A_L, A_R)$, so $\|\mu_1^*\|^2 = \|A_L\|^2 + \|A_R\|^2$. ∎

Of course we cannot ignore the effect of points in $S_j, j > 1$, on $\mu_1^{(1)}$. Accommodating these is straightforward.

**Lemma 9** *Choose any $\beta \in (0, l)$. Assume $\min(c, c^2)n^{1/2-\alpha} \geq 14$, $\min(n, c^2 n) \geq 18 + 8 \ln n$, $c^2 n \geq 512(\beta+1) \ln m$, $m \geq \max(4l^2, 2^{18}c^{-4})$. Then with probability at least $1 - l(m^{-\beta} + e^{-n/8})$, for each center-estimate $\mu_{i'}^{(1)} \in S_i$ with mixing weight more than $w_T$,*

$$\|\mu_{i'}^{(1)} - \mu_i\| \leq \tfrac{1}{4}c\sigma\sqrt{n}.$$

In other words, to get reasonably accurate estimates in the first round, we set $l = O(\frac{1}{w_{min}} \log k)$, and we need $c \gg n^{-1/4}$, $m \geq \max(4l^2, O(c^{-4}))$ and $c^2 n \geq \log \frac{1}{w_{min}}$.

## 6 Pruning

At the end of the first round of EM, let $C_j$ denote the center-estimates originally from $S_j$ which have high mixing weight, that is, $C_j = \{\mu_i^{(1)} : \mu_i^{(0)} \in S_j, w_i^{(1)} \geq w_T\}$. A simple clustering heuristic due to Hochbaum and Shmoys (1985), described in Section 3.3, is used to choose $k$ points from $\cup_j C_j$.

**Lemma 10** *If $c^2 n \geq 8 \ln 12l$ and $m \geq 40l$ then the sets $C_j$ obey the following properties.*

*(a) Each $C_j$ is non-empty.*

*(b) There is a real value $\Delta > 0$ such that if $x \in C_i$ and $y, z \in C_j$ ($i \neq j$) then $\|y - z\| \leq \Delta$ and $\|x - y\| > \Delta$.*

*(c) The pruning procedure identifies exactly one member of each $C_j$.*

*Proof.* (a) From Corollary 4 and Lemma 5 we already know that $|S_i| \geq \frac{3}{4}mw_i$, and that at most $\frac{5}{4}lw_i$ initial center-estimates are chosen from $S_i$. It was seen in Lemma 9 that each point in $S_i$ gives weight at least $1 - le^{-c^2 n/8}$ to center-estimates from $S_i$. It follows that at the end of the first round of EM, at least one of these center-estimates must have mixing weight at least

$$\frac{(\tfrac{3}{4}mw_i)(1 - le^{-c^2 n/8})}{m \cdot \tfrac{5}{4}lw_i} = \frac{3}{5l} \cdot (1 - le^{-c^2 n/8}) \geq w_T$$

(under the conditions on $m, l$), and therefore $C_i$ cannot be empty.

(b) Pick $x \in C_i$ and $y, z \in C_j$ for any pair $i \neq j$. Then $\|y - z\| \leq \Delta$ and $\|x - y\| \geq c_{ij}\sigma\sqrt{n} - \Delta$ where $\Delta$ is twice the precision of the center-estimates after the first round of EM. By the results of the previous section we may set $\Delta = \tfrac{1}{2}c\sigma\sqrt{n}$.

(c) There are $k$ true clusters and the pruning procedure picks exactly $k$ center-estimates. It will not pick two from the same true cluster because these must be at distance $\leq \Delta$ from each other, whereas there must be some untouched cluster containing a center-estimate at distance $> \Delta$ from all points selected thus far. ∎

## 7 The second round of EM

We now have one center-estimate $\tilde{\mu}_i^{(1)}$ per true cluster (for convenience permute their labels to match the $S_i$), each with mixing weight $\tfrac{1}{k}$ and covariance $\tilde{\sigma}^{(1)2} I_n$, where $\tilde{\sigma}^{(1)} = \sigma^{(0)}$. Furthermore each $\tilde{\mu}_i^{(1)}$ is within distance $\tfrac{1}{4}c\sigma\sqrt{n}$ of the corresponding true Gaussian center $\mu_i$. Such favorable circumstances will make it easy to show that the subsequent round of EM will achieve near-perfect clustering. The details are similar to those of the first round of EM and are omitted from this abstract. Combining the various results so far gives Theorem 1.

We can also bound the final mixing weights and variance. Here is an example.

**Lemma 11** *To the results of Theorem 1 it can be added that for any $i$,*

$$\frac{|S_i|}{m} \cdot (1 - ke^{-c^2 n/8}) \leq w_i^{(2)} \leq \frac{|S_i|}{m} + e^{-c^2 n/8}$$

## 8 The case of different spherical covariance matrices

A few changes need to be made when the data is drawn from a mixture $w_1 N(\mu_1, \sigma_1^2 I_n) + \cdots + w_k N(\mu_k, \sigma_k^2 I_n)$ in which the $\sigma_i$ might not be identical. In the algorithm itself, there are two changes.

**Initialization** Pick initial centers and mixing weights as before. For initial estimates of the variances use

$$\sigma_i^{\langle 0 \rangle 2} = \frac{1}{2n} \min_{j \neq i} \|\mu_i^{\langle 0 \rangle} - \mu_j^{\langle 0 \rangle}\|^2.$$

**EM** Run one round of EM, as before, to get the modified estimates $\mu_i^{\langle 1 \rangle}, \sigma^{\langle 1 \rangle}, w_i^{\langle 1 \rangle}$.

**Pruning** Again remove center-estimates with weight below $w_T$. The only difference in the remainder of the pruning procedure is that the distance between centers $\mu_i^{\langle 1 \rangle}$ and $\mu_j^{\langle 1 \rangle}$ is now weighted by the individual variances,

$$d(\mu_i^{\langle 1 \rangle}, \mu_j^{\langle 1 \rangle}) = \frac{\|\mu_i^{\langle 1 \rangle} - \mu_j^{\langle 1 \rangle}\|}{\sigma_i^{\langle 0 \rangle} + \sigma_j^{\langle 0 \rangle}}.$$

**EM** One last step of EM, as before.

The modified distance measure in the pruning step is meant, roughly, to compensate for the fact that part of the distance between $\mu_i^{\langle t \rangle}$ and $\mu_j^{\langle t \rangle}$ is on a scale of $\sigma_i^{\langle t \rangle}$ while part of it is on a scale of $\sigma_j^{\langle t \rangle}$. The analysis follows roughly the same outline as before, with a few extra subtleties. An additional assumption is needed,

$$c_{ij}^2 \max(\sigma_i^2, \sigma_j^2) \geq |\sigma_i^2 - \sigma_j^2| \text{ for all } i, j,$$

in order to rule out situations in which one cluster is nested within another. The final theorem remains the same, the error $\|\mu_i^{\langle 2 \rangle} - \mu_i\|$ now being proportional to $\sigma_i$ instead of to the common $\sigma$ of the previous case.

## 9  Concluding remarks

This paper provides principled answers to many questions surrounding EM: how many clusters should be used, how the parameters ought to be initialized, and how pruning should be carried out. Some of the intuition presented here confirms current practice; some of it is new. Either way, this material should be of interest to practitioners of EM.

But what about the claim that EM can be made to work in just two rounds? This requires what we call the

**Strong Gaussian assumption.** The data are i.i.d. samples from a true mixture of Gaussians.

This assumption is the standard setting for other theoretical results about EM, but is it reasonable to expect of real data sets? We recommend instead the

**Weak Gaussian assumption.** The data looks like it comes from a mixture of Gaussians in the following sense: for any sphere in $\mathbb{R}^n$, the fraction of the data that falls in the sphere is the expected fraction under the mixture distribution, $\pm\epsilon_0$, where $\epsilon_0$ is some term corresponding to sampling error and will typically be proportional to $m^{-1/2}$, where $m$ is the number of samples. Some other concept class of low VC dimension can be substituted for spheres.

The strong assumption immediately implies the weak assumption (with high probability) by a large deviation bound, since the concept class of spheres in $\mathbb{R}^n$ has small VC dimension. What kinds of conclusions follow from the strong assumption but not the weak one? Here is an example: "if two data points are drawn from $N(0, I_n)$ then with overwhelming probability they are separated by a distance of at least $\sqrt{n}$". The weak assumption does not support this; with just two samples, in fact, the sampling error is so high that it does not allow us to draw any non-trivial conclusions at all.

It is often argued that the Gaussian is the most natural model of a cluster because of the central limit theorem. However, central limit theorems, specifically Berry-Esséen theorems (Feller, 1966), yield Gaussians in the sense of the weak assumption, not the strong one. For the same reason, the weak Gaussian assumption arises naturally when we take random projections of mixtures of product distributions (Diaconis and Freedman, 1984). Ideally therefore, we could provide performance guarantees for EM under just this condition. Perhaps our analysis can be extended appropriately. For an example of what needs to be changed in the algorithm, consider that the weak assumption allows $\sqrt{m}$ out of $m$ data points to be placed arbitrarily. An outlier removal procedure might be necessary to prevent EM from being confused by this possibly malicious noise.